\documentclass[runningheads]{llncs}

\usepackage[T1]{fontenc}

\usepackage{graphicx}
\usepackage{float}

\begin{document}

\raggedbottom
\title{Towards VLA-Dreamer: Refining VLA Behavior Using World Models}

\author{Parsa Mastouri Kashani \and Jan-Gerrit Habekost \and Stefan Wermter}

\authorrunning{P. Mastouri Kashani et al.}

\institute{Knowledge Technology, Dept. of Informatics, University of Hamburg, Germany\\
\email{\{parsa.mastouri.kashani, jan-gerrit.habekost, stefan.wermter\}@uni-hamburg.de}}

\maketitle

\begin{abstract}
Vision-Language-Action models (VLAs), while showing strong potential for robot control, require massive amounts of high-quality imitation learning data. Moreover, the absence of an explicit world model casts further doubt on their control capabilities. In this concept paper, we propose a novel architecture that addresses sample efficiency in VLAs by training a predictive world model on the embedding space of the VLA's vision encoder. We hypothesize that these embeddings are action-relevant and usable for future prediction. To this end, we propose using the suggested architecture to investigate how well these embeddings predict the future based on actions, as the inability to do so would mark a key limitation of VLA architectures: the lack of a non-lossy implicit world model to simulate real-world dynamics. The proposed architecture differs from the standard world model dynamics as the loss comes from the embedding space rather than the pixel space, similar to joint embedding predictive architectures. Furthermore, the trained world model can be utilized for short-term planning tasks by sampling VLA actions given goal images. We intend to examine the richness of vision embeddings in VLAs and reduce their high data requirements through a world model that can also generate plans during inference.

\end{abstract}

\section{Introduction}

Robotics has seen a recent surge with the introduction of Vision-Language-Action models (VLAs) \cite{BBC23,KPKX24,OGWP24}, which are, in essence, Vision-Language Models (VLMs) that are themselves extensions of LLMs equipped with a vision backbone. However, VLAs remain reactive models with no explicit model of the real world. World-Action Models (WAMs) \cite{WJCC24,YGZG26} address this by using video-generator backbones that can predict how the world might evolve. The JEPA \cite{LeC22} line of work, however, argues that because these generative objectives lie in pixel space, the models are forced to ``think in pixels'', which is not an optimal surrogate for a mental model of the world, since such a model must account for every detail of a scene rather than reasoning at the right level of abstraction. This forces the model to attend to unnecessary details, such as the leaves on the sidewalk in a self-driving car application.

In response, models such as V-JEPA 2-AC \cite{ABF25} embed observations in a rich embedding space free of the aforementioned pixel-space overhead. Their vision encoder is first trained with a mask-denoising objective, which forces the encoder to retain prediction-relevant information. An action-conditioned predictor is then trained on top of the frozen encoder to predict the next state given the current observations and an action. However, they rely on Model Predictive Control (MPC), an expensive planning procedure that iteratively selects actions whose predicted future embeddings minimize the distance to the goal embedding. Furthermore, their solution requires goal images, or sub-goal frames, of the expected outcome at inference time, which might not be available and which limit the horizon of the tasks the model can perform between consecutive sub-goals.

An alternative would be to train a lightweight MLP-based policy with RL, trained from scratch inside that embedding space, to achieve the same goal at a lower computational cost. Here, the world model serves as a simulator, while the agent remains small and simply reacts to the given state. However, such models lack the semantic priors needed for language conditioning and are forced to remain small to be trainable using RL without overfitting. Furthermore, designing a reward function for such a system is challenging as the objective changes from one task to another.

We propose a novel approach: reusing the frozen vision encoder of a VLA, we train a world model that predicts the next observation embedding given an action, the last few observation embeddings, and the robot's proprioception. We then continue training our VLA, which has already been fine-tuned on our embodiment, inside this world model using reinforcement learning.

This approach addresses sample efficiency, and we hypothesize that having the VLA explore outcomes inside the world model would induce an efficient inner world model within the VLA agent itself, as it would be aware of the consequences of its actions. Furthermore, using a VLA allows us to gain further insight into the policy through question answering in the embedding space, since the embeddings are similar to the language-aligned vision encoder embeddings.

\section{Methodology}

\subsection{Training and Choice of VLA}
\subsubsection{Choice of VLA}
Many VLAs, such as SmolVLA \cite{SACK25}, keep the vision encoder of their VLM backbone frozen during training. Such decisions, while plausible for training on limited data domains, would rely on the vision encoder to have encoded all necessary action-related information from the image, such as the precise location of each object in the scene. Although the encoder in such models is trained on VQA-style tasks during VLM pre-training, it might not capture all the action-related features needed by the policy. As a result, we pivot to using a VLA whose training has also updated the vision encoder. We hypothesize it would lead to better action-related features for future embedding generation.

Based on these considerations, we propose to build on $\pi_0$-FAST \cite{PSID25} and OpenVLA \cite{KPKX24} as our VLAs of choice, since they both operate on token probabilities rather than diffusion or flow-matching, which allows for a simpler RL pipeline. Furthermore, DINOv2 \cite{ODM24} has already proven to be able to generate future embeddings in DINO-WM \cite{ZPLP24}, and OpenVLA makes use of a fusion of DINOv2 and SigLIP \cite{ZMK23} adapted from the Prismatic line of VLMs \cite{karamcheti2024prismatic}.

\begin{figure}
\vspace{-15pt}
  \centering
  \includegraphics[width=0.9\textwidth]{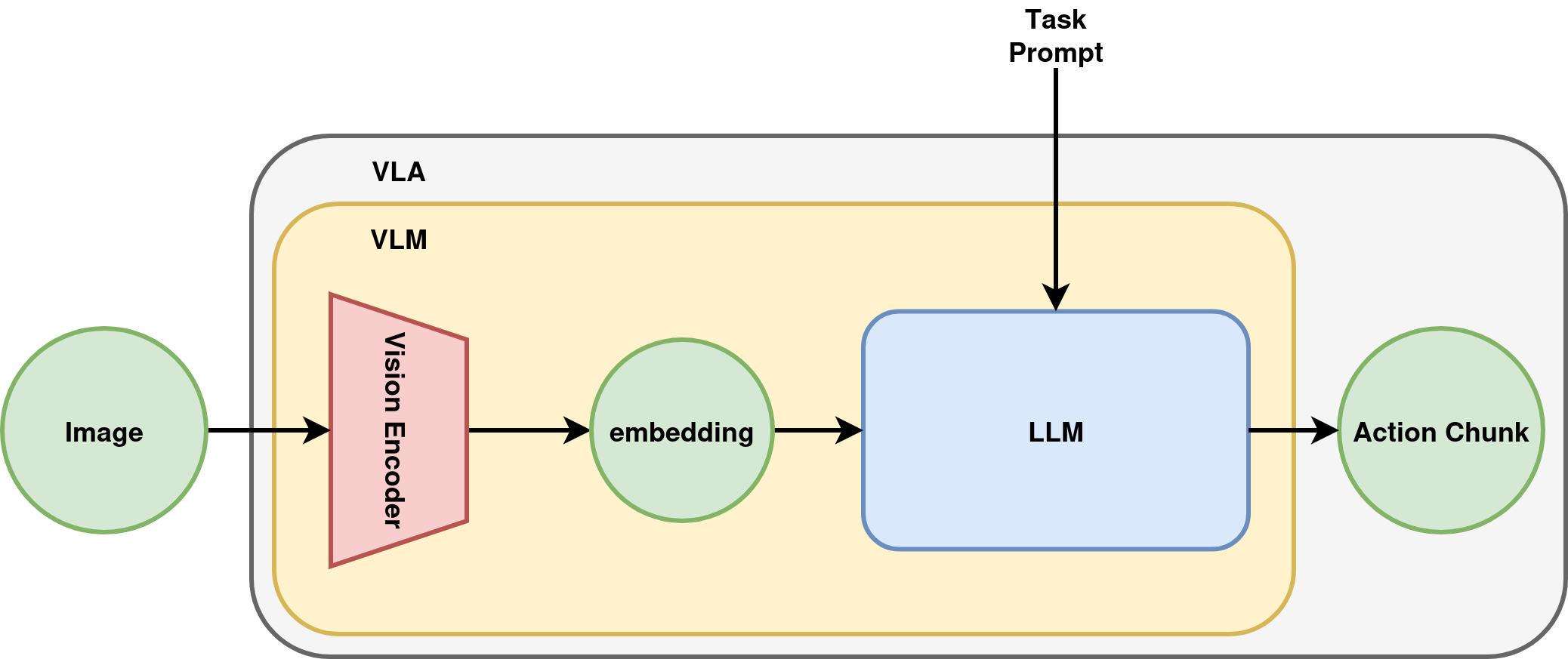}
  \caption{We fine-tune our pre-trained VLA model on our NICOL robot's imitation learning dataset before further training the model using RL.}
  \label{fig:watermark}
  \vspace{-25pt}
\end{figure}

\subsubsection{VLA Training}
First, we fine-tune the VLAs given our current imitation learning dataset for NICOL \cite{KASFHEW23}, which consists of language commands and teleoperation data, including proprioception (joint positions) and the camera frames of the robot at each timestep. To avoid misalignment between the world model embeddings and the VLA's vision encoder, we freeze the vision encoder. To avoid losing semantic knowledge, we also apply strong regularization and freeze most weights in the LLM backbone of the VLA.

\subsection{World Model}

To train the world model we use a mixture of random actions and demonstrations by semi-expert or expert policies controlling the robot. This ensures good coverage of the world dynamics as training on expert demonstrations or random actions alone might leave the robot environment unexplored. 
The world model for $\pi_0$-FAST would be in charge of predicting the next proprioceptive state and the next-timestep frame embeddings conditioned on the current action, the current robot proprioceptive state, and a history of recent camera observations.
The world model for OpenVLA would also predict proprioception, but only as an auxiliary loss, as proprioception is not a necessary part of the OpenVLA training recipe and would not be fed to the VLA model during RL training.
Alongside the world model, we train an image decoder, but only for probing and further visualization, as the embeddings output by the world model are not optimal for image reconstruction, and the decoder is only used as a probe to see a lossy picture of what the model is dreaming.

\begin{figure}
  \centering
  \includegraphics[width=1.0\textwidth]{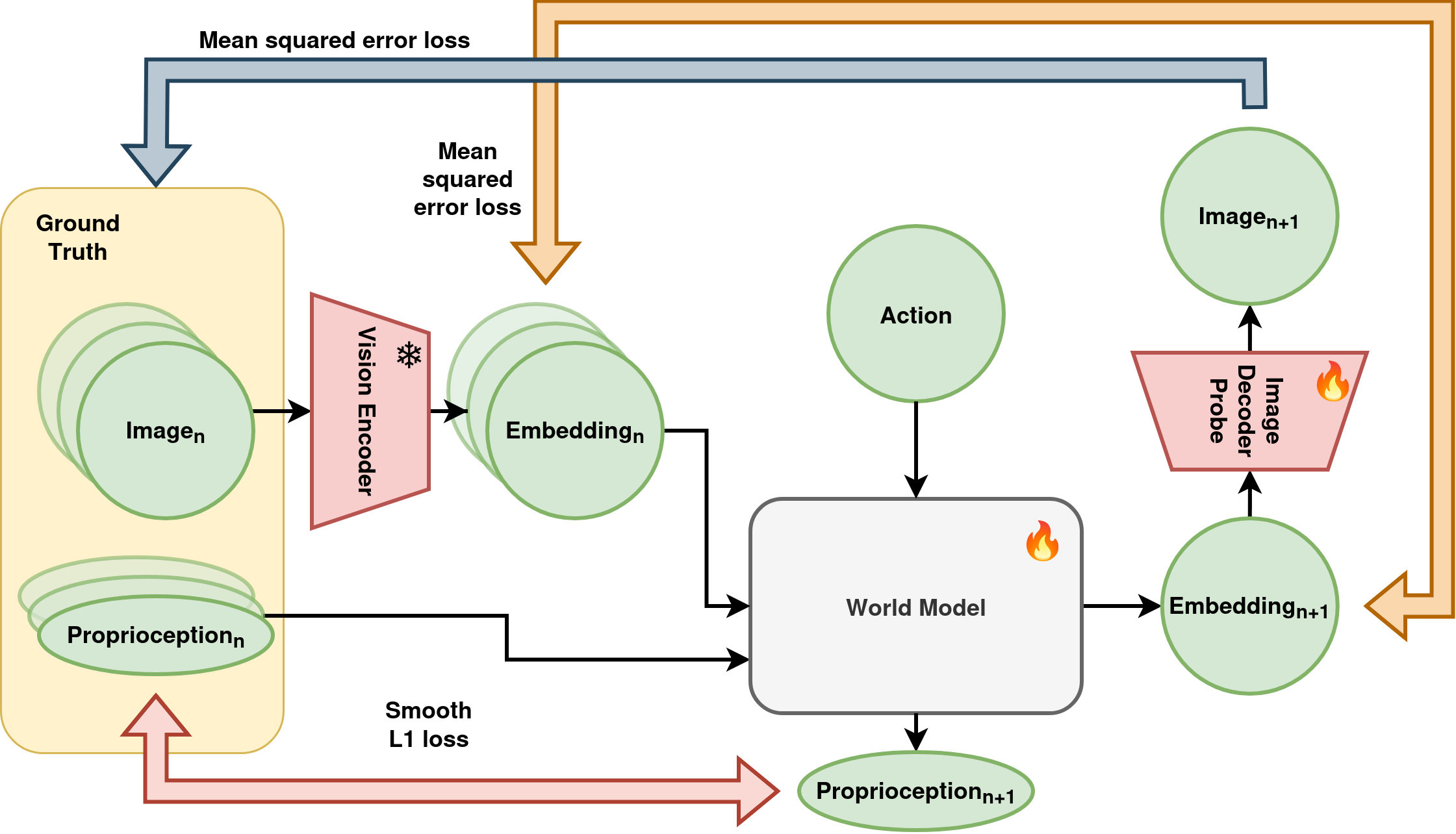}
  \caption{This figure demonstrates how the world model gets trained. A history of embeddings and proprioception, along with an action, is input to the world model, which predicts the next observation embedding.}
  \label{fig:WorldModel}
\end{figure}

A summary of the process can be seen in Figure~\ref{fig:WorldModel}.

\subsection{Embedding Assessment}

To test our central hypothesis that the frozen vision encoder of a VLA model retains enough action-relevant information to allow future prediction, we attach two probes to the vision encoder alongside the image decoder probe: a semantic segmentation probe and a depth estimation probe. We hypothesize that a capable manipulation policy is aware of object masks and scene geometry, and as a result, strong probe performance provides evidence that the embeddings encode them.
Weak performance on the two given tasks indicates a limitation for both action generation and future prediction. The two quantitative values would be mIoU for segmentation and RMSE with threshold accuracy for depth.

We expect the embeddings from OpenVLA's Prismatic backbone to already perform well at depth estimation, because of its use of DINOv2. By contrast, the embeddings from $\pi_0$-FAST may be less capable in this regard as its vision encoder begins as a SigLIP model and then undergoes VLM training as part of PaliGemma-3B \cite{BSS24}, which is further trained under the $\pi_0$-FAST robot training recipe.
This comparison therefore offers a useful demonstration of what additional features the vision encoder acquires during VLA pre-training.

Unlike models such as OpenVLA, the $\pi$ family of VLAs \cite{BBD25} are not restricted to a single input image. Depth estimation is therefore not constrained to the monocular setting, which can make the task easier for these models.
Finally, the same probes apply to the embeddings \emph{predicted} by the world model. Beyond visualization, this quantifies whether imagined futures preserve object masks and depth as the imagination horizon grows, turning probe accuracy on predicted embeddings into a direct measure of world-model fidelity and drift.

\subsection{RL Training}
After fine-tuning the VLA on our embodiment and training the world model to predict future embeddings, we pick video frames of the robot executing tasks, for example, stacking a square block onto another. First, we embed the initial table configuration and then the goal image. Now we provide the correct prompt to our VLA policy and use the world model as a simulator for our policy to interact with.
The negative mean squared distance between the current state embedding and the goal state embedding would be used as a dense reward function for the RL policy. 
The choice of VLAs, which output token probabilities, allows us to sample actions and to run model-free RL algorithms such as PPO.
Since the VLA is already competent, RL serves to refine fine-grained actions rather than discover them, relaxing reward-shaping demands. We restrict the imagination horizon and keep the policy near its pre-trained prior to limit both compounding drift and exploitation of world model error.

\begin{figure}
  \centering
  \includegraphics[width=0.9
  \textwidth]{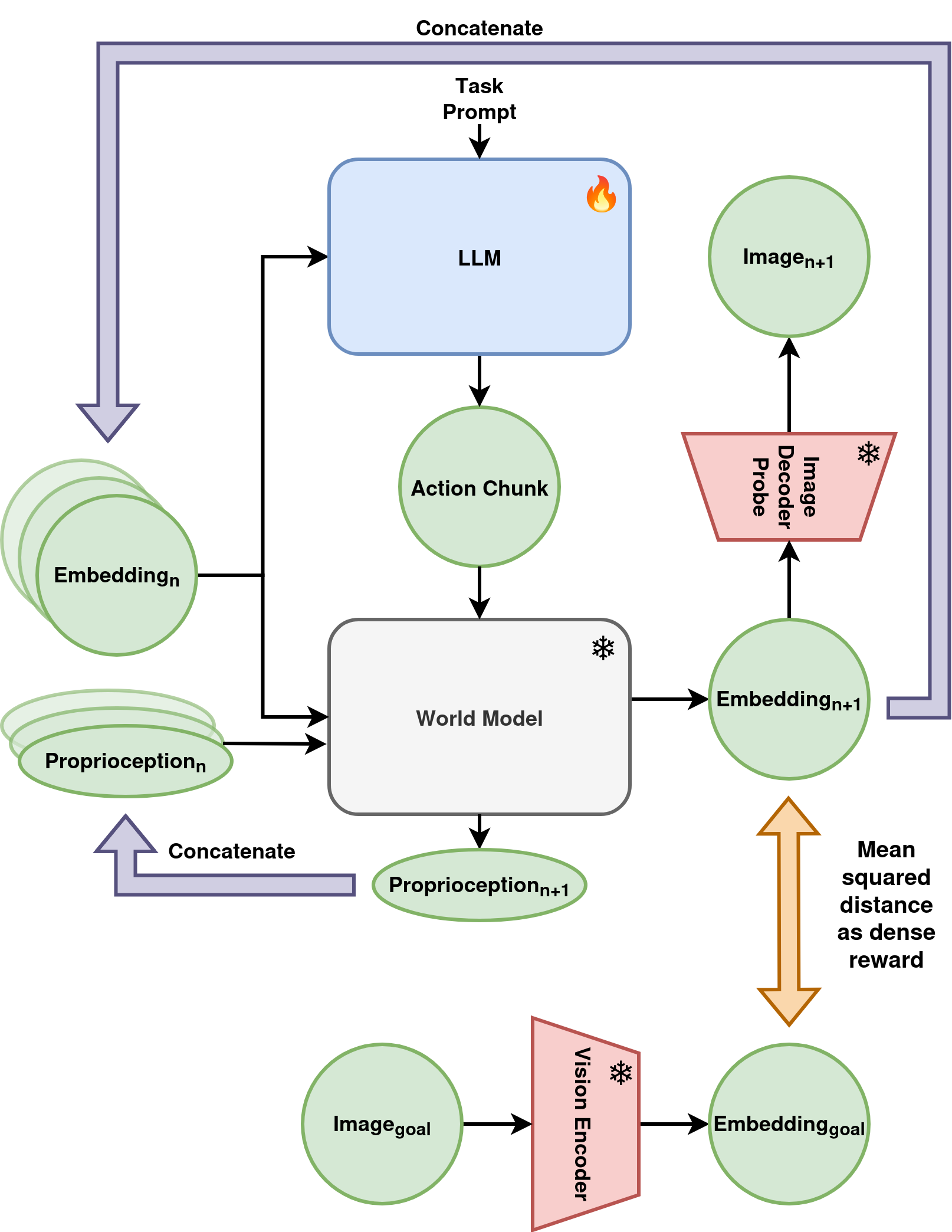}
  \caption{After training the VLA on our embodiment's teleoperation dataset and training our world model and freezing it, we proceed to use the presented model for RL. By using the negative mean squared distance between the current state embedding and the goal state embedding, we provide a dense reward function for the VLA to act inside the world model.
  }
  \label{fig:RL}
\end{figure}

A summary of our approach can be seen in Figure~\ref{fig:RL}.

\section{Discussion and Challenges}
The major challenge for the system is its reliance on the vision encoder in VLA models to produce next-observation embeddings. Even though VLA models rely on their vision encoder to function and the embeddings are rich enough for action generation, they might not be enough for future prediction. While using multiple observations and camera viewpoints for embedding generation, combined with proprioception data, helps mitigate this, it might not be enough in highly detailed and cluttered scenarios.

Another concern is the effect of drift and compounding errors in the world model predictions. The amount of drift indicates the horizon of the imagination. Substantial drift might only allow us to implement an RL policy to refine the fine-grained actions rather than plan over longer trajectories.

Our approach can be a better alternative to using simulation for generating trajectories. Generating imagination requires less compute than building and running a full simulator, and it avoids the sim-to-real gap as the world model is trained on real demonstrations and therefore captures task-relevant aspects of the environment that hand-built simulations often lack, especially in manipulation scenarios. However, as stated earlier, the performance depends on the quality of the embeddings coming from the vision encoder, and training an adequate world model requires substantial compute.

Finally, using image goals to produce embeddings for the RL reward might prove difficult, as there may be a multitude of ways to capture visual snippets of the goal state.
Furthermore, the mean squared distance between embeddings, depending on the task, might not be monotonic and hence might pose a challenge.
Our hypothesis is that the semantic abstractions of the embeddings and the language-defined goal given by the prompt would mitigate this phenomenon. Moreover, the VLA is under heavy regularization and includes most of the needed knowledge inside its pre-training recipe, so the reward would help with refining the trajectory.
If proven insufficient, the reward can be replaced or augmented with a sparse success signal from a classifier for task success.

\section{Conclusion}
We proposed an architecture that trains a predictive world model using the embeddings of a VLA's frozen, language-aligned vision encoder, and that refines the fine-tuned VLA through reinforcement learning inside this model. Because the policy input, the prediction target, and the goal-distance reward all share one representation, a single encoder supports action generation, future prediction, and goal specification.
Beyond the architecture itself, the proposed experiments test whether the embeddings a VLA already relies on for action generation are rich and predictable enough to serve as the latent space of a world model. A positive result would suggest VLAs carry an implicit, usable model of world dynamics, while a negative one would point to a limitation of current architectures, namely the absence of a non-lossy implicit world model. Our next step is to train the world model and probes, enabling us to measure embedding-prediction errors and probe fidelity as the imagination horizon grows, quantifying whether imagined futures stay geometrically coherent enough to support planning and RL.

\begin{credits}
\subsubsection{\ackname} 
The authors gratefully acknowledge support from the German Research Foundation DFG under project LUMO (No 551629603) and funding from Horizon Europe under the MSCA grant agreements No 101226624 (GREET), No 101168792 (SWEET), and No 101072488 (TRAIL).

\subsubsection{\discintname}
The authors have no competing interests to declare that are relevant to the content of this article.
\end{credits}

\end{document}